\documentclass{article}

\usepackage{PRIMEarxiv}

\usepackage[utf8]{inputenc} 
\usepackage[T1]{fontenc}    
\usepackage{hyperref}       
\usepackage{url}            
\usepackage{booktabs}       
\usepackage{amsmath}
\usepackage{amsfonts}       
\usepackage{nicefrac}       
\usepackage{microtype}      
\usepackage{lipsum}
\usepackage{fancyhdr}       
\usepackage{graphicx}       
\usepackage{subcaption}
\usepackage{tabularx}
\graphicspath{{media/}}     

\title{Multi-Objective Decision Transformers for Offline Reinforcement Learning
}

\author{%
  Abdelghani Ghanem\\
  TICLab, College of Engineering and Architecture\\
  International University of Rabat\\
  Morocco\\
  LTCI, Communications and Electronics Department\\
  Telecom ParisTech, Institut Polytechnique de Paris\\
  France\\
  \texttt{abdelghani.ghanem@uir.ac.ma} \\
  \And
  Philippe Ciblat \\
  LTCI, Communications and Electronics Department\\
  Telecom ParisTech, Institut Polytechnique de Paris\\
  France\\
  \texttt{philippe.ciblat@telecom-paris.fr} \\
  \And
  Mounir Ghogho \\
  TICLab, College of Engineering and Architecture\\
  International University of Rabat\\
  Morocco\\
  \texttt{mounir.ghogho@ac.uir.ma} \\
}

\begin{document}
\maketitle

\begin{abstract}
Offline Reinforcement Learning (RL) is structured to derive policies from static trajectory data without requiring real-time environment interactions. Recent studies have shown the feasibility of framing offline RL as a sequence modeling task, where the sole aim is to predict actions based on prior context using the transformer architecture. However, the limitation of this single task learning approach is its potential to undermine the transformer model's attention mechanism, which should ideally allocate varying attention weights across different tokens in the input context for optimal prediction. To address this, we reformulate offline RL as a multi-objective optimization problem, where the prediction is extended to states and returns. We also highlight a potential flaw in the trajectory representation used for sequence modeling, which could generate inaccuracies when modeling the state and return distributions. This is due to the non-smoothness of the action distribution within the trajectory dictated by the behavioral policy. To mitigate this issue, we introduce action space regions to the trajectory representation. Our experiments on D4RL benchmark locomotion tasks reveal that our propositions allow for more effective utilization of the attention mechanism in the transformer model, resulting in performance that either matches or outperforms current state-of-the art methods.
\end{abstract}

\section{Introduction}

\begin{figure}[t]
    \centering
    \begin{subfigure}[b]{0.27\textwidth}
        \centering
        \includegraphics[width=\textwidth]{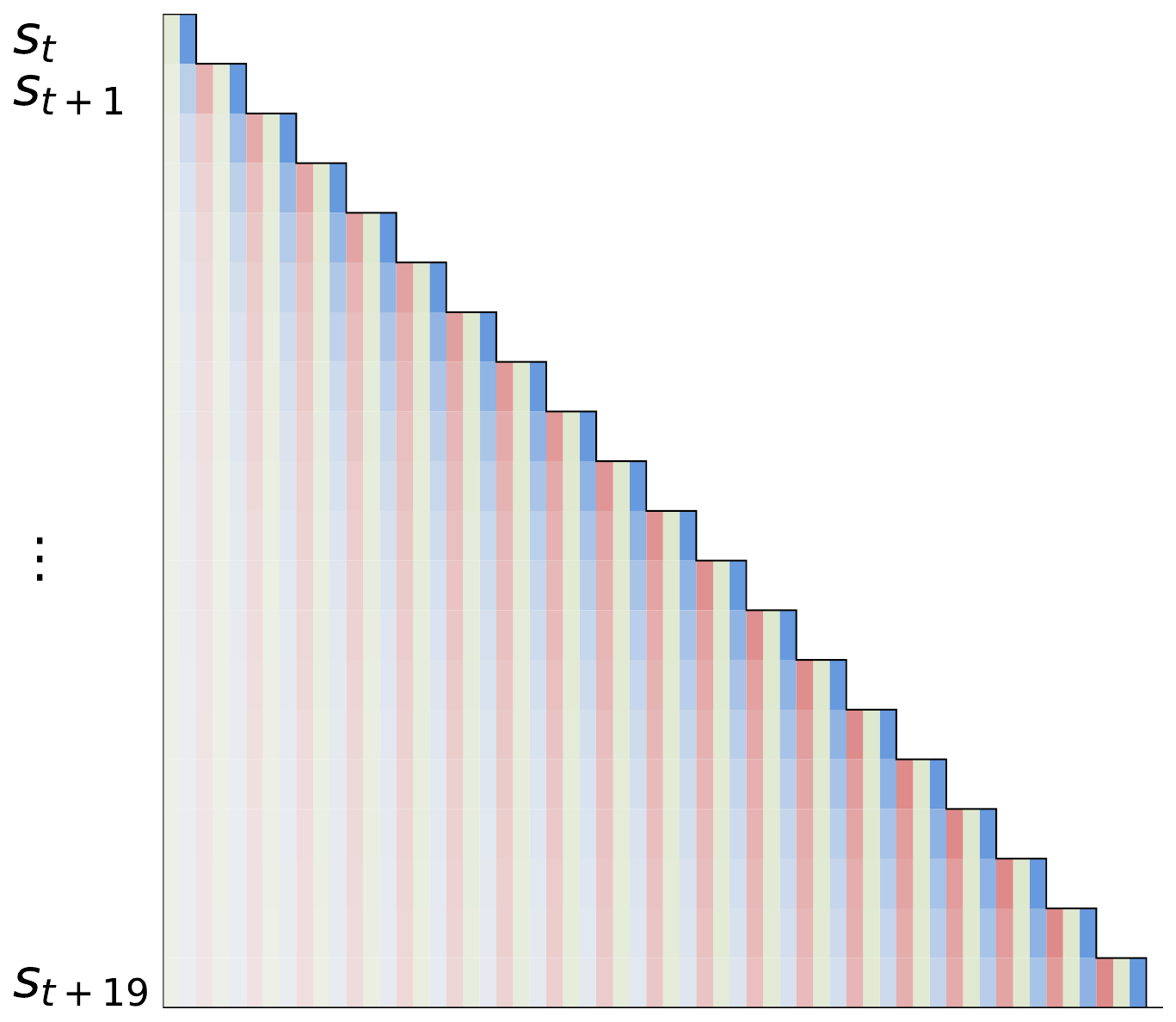}
        \label{fig:subfig1}
    \end{subfigure}
    \hfill
    \begin{subfigure}[b]{0.23\textwidth}
        \centering
        \includegraphics[width=\textwidth]{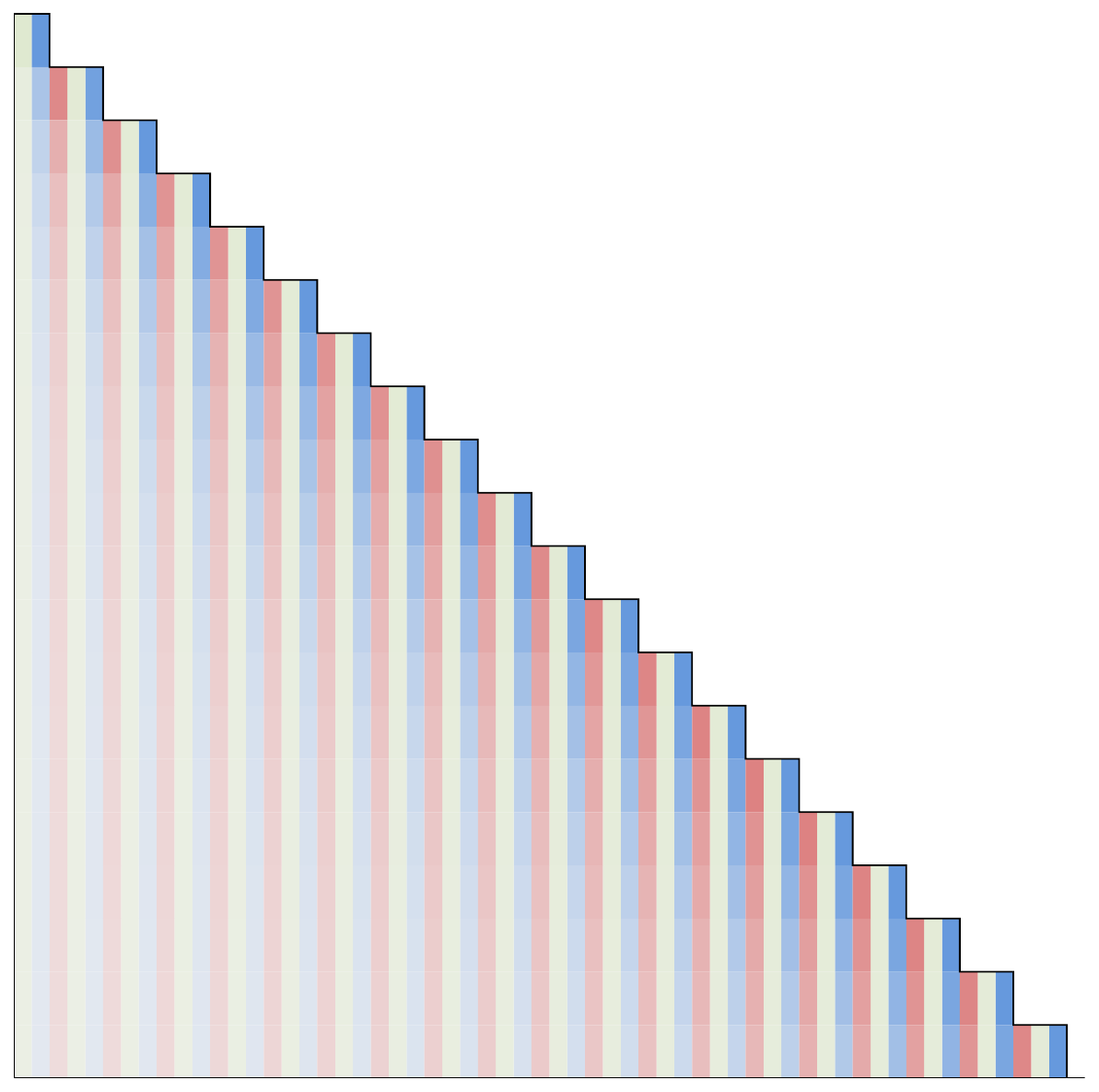}
        \label{fig:subfig2}
    \end{subfigure}
    \hfill
    \begin{subfigure}[b]{0.23\textwidth}
        \centering
        \includegraphics[width=\textwidth]{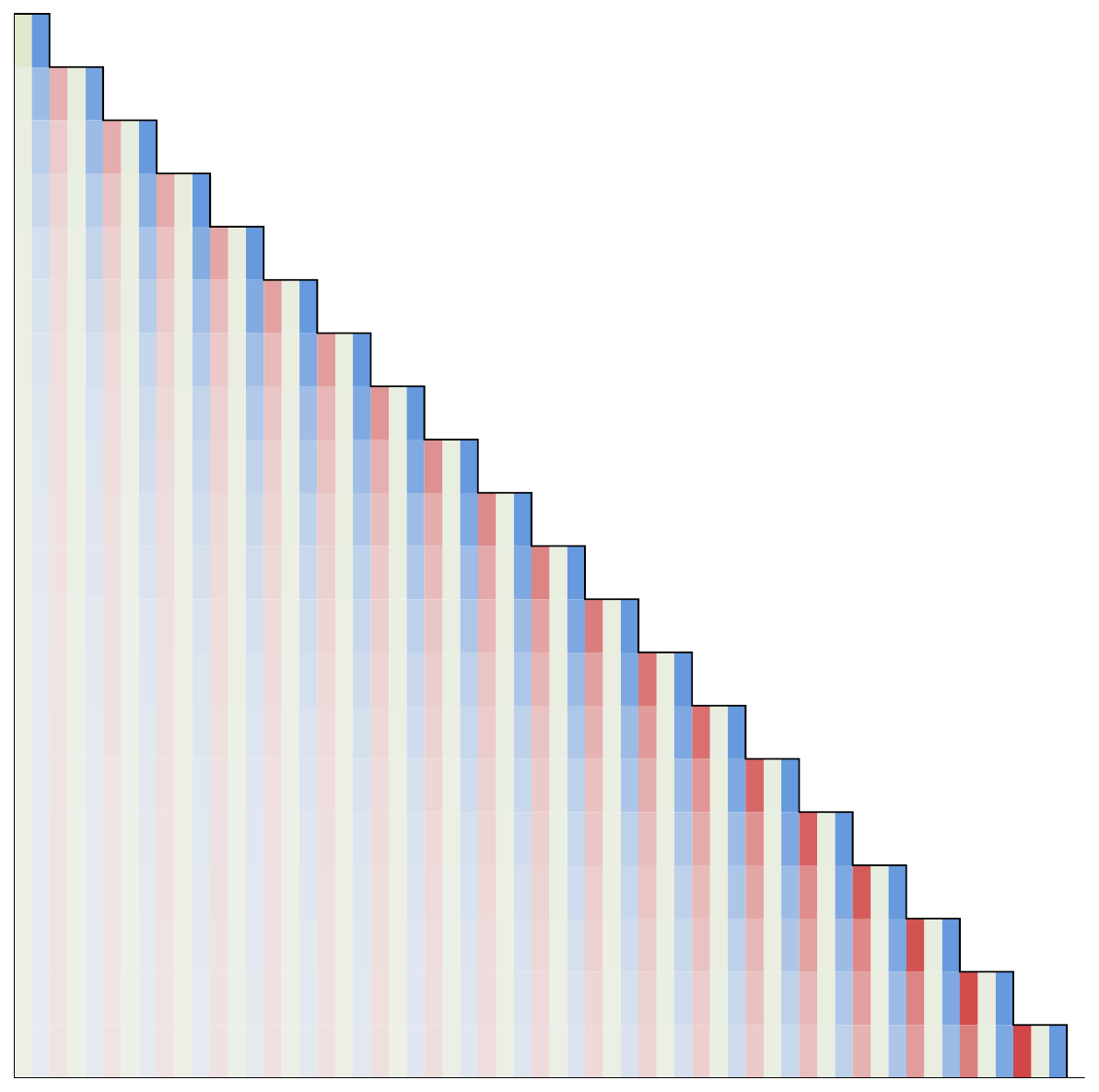}
        \label{fig:subfig3}
    \end{subfigure}
    \hfill
    \begin{subfigure}[b]{0.23\textwidth}
        \centering
        \includegraphics[width=\textwidth]{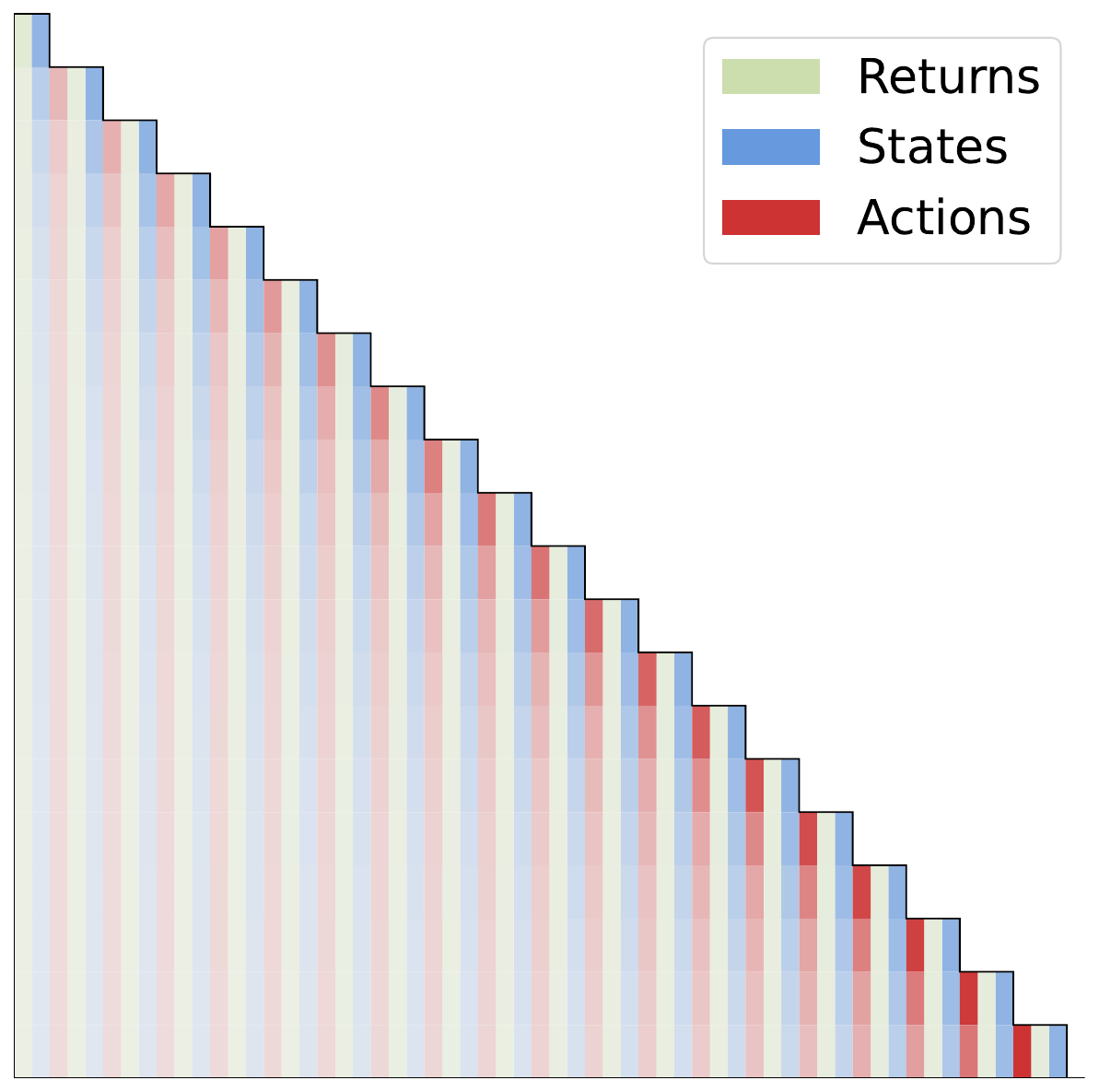}
        \label{fig:subfig4}
    \end{subfigure}
    \caption{Attention patterns per block for DT during action prediction on the walker2d environment}
    \label{fig:attention_maps_dt}
\end{figure}

In the realm of offline Reinforcement Learning (RL), traditionally referred to as batch RL \cite{fujimoto2019offpolicy, offline_rl_2}, the objective is for an agent to acquire effective policies solely from static, finite datasets. These datasets are often harvested by an arbitrary, possibly unknown process, devoid of online interaction. This paradigm is particularly appealing across a range of real-world applications where data is readily available, but exploratory actions using untrained policies are prohibitive due to high costs or potential risks. Examples of such domains include robotics, recommender systems, education, autonomous driving, and healthcare. \\
A compelling approach to offline RL, primarily thanks to its algorithmic elegance, interprets the problem as a sequence modeling task \cite{chen2021decision, tt}. This approach employs strategies akin to those used in large-scale language modeling. The key insight here is that an experience deemed suboptimal for one task could be optimal for another. By conditioning on particular parameters like desired returns or using simple planning heuristics like beam search, such experiences can be harnessed for direct behavior cloning \cite{rvs}. The Decision Transformer (DT) \cite{chen2021decision}, in particular, stands out for modeling sequences of three types of tokens: returns, states, and actions, using the transformer architecture. The DT is trained exclusively for action prediction, while other tokens are incorporated into the trajectory to guide action prediction during inference. However, this narrow objective, focused solely on action prediction, may not fully exploit the transformer architecture's potential. As illustrated in Figure \ref{fig:attention_maps_dt}, which presents the attention patterns of the DT model introduced in \cite{zheng2022online}, the model pays close attention to all context tokens during action prediction. Intriguingly, it appears to assign similar weightage to tokens of the same type, and distinct attention blocks seem to learn similar aspects. This suggests that the DT model might not be making effective use of the transformer's attention mechanism. This observation is in line with the counter-intuitive finding recently reported in \cite{rvs}, which demonstrated that a simple two-layers MLP, which takes only the previous state and a desired return as input, can achieve performance on par with DT.

Motivated by the above-mentioned observation, this paper commences with an exploration of the potential implications of incorporating return and state prediction alongside the original action prediction within the DT model. This investigation is conducted within the framework of multi-objective optimization, thereby casting the sequence modeling problem in a new light. Although it has previously been suggested that incorporating state and return prediction does not significantly enhance the efficacy of the DT model \cite{chen2021decision}, we here demonstrate that through a judicious selection of the model architecture, such as the inclusion of multiple attention heads, and an appropriate probabilistic framework that employs Gaussian prediction heads for states and returns, the attention mechanism intrinsic to the transformer model can be more effectively utilized. As a result, our model, which we coin Multi-Objective Decision Transformer (\textbf{MO-DT}), demonstrates a level of performance that is substantially higher than that of vanilla DT and is competitive with current state-of-the art techniques.

Following the analysis of Figure \ref{fig:attention_maps_dt}, our second observation pertains to the DT model's noticeable reliance on antecedent actions. We posit that the trajectory-centric representation in the DT framework could inadvertently tether the learned policy to the behavior policy, given the conditioning on preceding actions within the trajectory. In the second part of our paper, we further explore the feasibility of constructing another multi-objective DT variant, namely the Trust Region Decision Transformer (\textbf{MO-TRDT}), that learns superior policies with diminished dependence on the behavioral policy. Our approach fundamentally hinges on modeling the conditional distributions of broader segments of the action space, that we refer to as 'action regions', conditioned on historical states, returns, and other prior information, in conjunction with the modeling of raw action distributions. This implies performing sequence modeling at a higher degree of abstraction for the behavioral policy. Experimental results corroborate our claim that this strategy shifts the attention of the MO-DT model more towards the previous action regions in the trajectory and away from raw actions during action prediction, while disregarding them during state and return prediction. This provides an enhanced flexibility for the learned policy, thus fostering improved generalization. Several experiments on the gym datasets from the D4RL benchmark\cite{fu2021d4rl}, shows that our two proposed approaches substantially outperforms the original DT, while matching or outperforming current state-of-the art methods.

In summary, in this work, we present two main contributions in the context of offline RL that expand upon the DT framework: (i) we introduce MO-DT, an enhanced DT variant that promotes focused attention through multi-objective optimization, encompassing return, state, and action predictions; our findings substantiate MO-DT's ability to allow for more effective use of the attention mechanism in the transformer model; (ii) we further extend the MO-DT model to another variant, MO-TRDT, which includes the modeling of the conditional distributions of broader segments of the action space. This higher level of abstraction in sequence modeling results in a learned policy with reduced reliance on the behavioral policy.



\section{Related Work}

In this section, we provide an overview of the related work relevant to MO-DT and MO-TRDT, focusing on three main fields: (1) transformers for reinforcement learning, (2) behavior regularized offline RL, and (3) regularization techniques in transformers.
\paragraph{Transformers for Reinforcement Learning.}
Recent advancements in casting reinforcement learning as a supervised learning problem have been noteworthy \cite{schmidhuber2019reinforcement, srivastava2021training}. As a further development of this paradigm, numerous studies have suggested treating offline RL as a context-conditioned sequence modeling problem \cite{chen2021decision, tt, zheng2022online}. This approach places greater emphasis on the predictive modeling of action sequences based on task specifications (e.g., returns or goals) rather than explicitly learning Q-functions or policy gradients, as commonly done in conventional model-free RL methods. \cite{chen2021decision} implemented a transformer \cite{vaswani2017attention, radford2018improving} as a model-free, context-conditioned deterministic policy, while \cite{zheng2022online} introduced a more general probabilistic formulation that employs a stochastic policy, which proved to be more effective in offline settings and can be fine-tuned efficiently in online scenarios. \cite{tt} employed a transformer as both a model and a policy, revealing that the integration of beam search can improve performance beyond several purely off-policy RL approaches. Later research demonstrated that the efficacy of these sequence modeling techniques could be enhanced through the adoption of more advanced training methodologies for sequence models, commonly used in the natural language processing literature, including transfer learning \cite{reid2022wikipedia}, self-training \cite{wang2022bootstrapped}, or contrastive learning \cite{konan2023contrastive}. In our work, we interrogate the foundational design choices of DT, as opposed to leveraging complex training methodologies. Naturally, the incorporation of these sophisticated techniques into our proposed models could pave the way for further research. Notably, given that our approach encompasses the modelling of both states and returns, an examination of the application of self-training scenarios to our model could yield insightful outcomes.

\paragraph*{Behavior Regularized Offline RL.}

Unlike sequence modeling approaches, off-policy reinforcement learning methods focus on learning an approximate value function, and improve the policy by optimizing this value function. The majority of off-policy RL methods can be applied in offline contexts; however, they often exhibit poor performance due to "extrapolation error." This type of error in policy evaluation arises when agents have difficulty accurately determining the value of state-action pairs that are absent in the dataset. Consequently, this issue negatively impacts policy improvement, as agents tend to select out-of-distribution actions with overestimated values, resulting in suboptimal performance \cite{fujimoto2019offpolicy}. As a result, many recently introduced offline RL approaches are adaptations of off-policy RL strategies that integrate constraints or regularizers to curtail deviations from the behavior policy. Various methods can be employed to incorporate constraints, such as appending a supervised learning term to the policy improvement objective \cite{fujimoto2021minimalist}, utilizing an explicit density model \cite{wu2019behavior, fujimoto2019offpolicy, kumar2019stabilizing}, or implementing implicit divergence constraints \cite{nair2021awac, wang2021critic, peters2007reinforcement, peng2019advantage}. In addition, regularizers can be directly integrated into the Q-function to assign lower values for out-of-distribution actions \cite{kostrikov2021offline2, kumar2020conservative}. Furthermore, \cite{kostrikov2021offline} have recently proposed an expectile-based implicit Q-learning algorithm that circumvents out-of-distribution actions by using an implicit constraint during policy extraction. In contrast, as our approach is founded on DT, which is inherently closely related to the data-generating process, we aim to permit the learned policy to deviate marginally from the behavior policy.

 \paragraph*{Regularization of transformer models.}

In the context of transformer models, various regularization approaches have been proposed with the goal of promoting diversity among the outputs of attention heads. This diversity is essential for effectively attending to information from different representation subspaces at distinct positions in the input context. Several techniques have been explored to achieve this objective, such as introducing an additional loss term to encourage diversity among attention head outputs \cite{li2018multihead}, leveraging dropout mechanisms \cite{JMLR:v15:srivastava14a} to foster diversity in attention heads by randomly masking attention heads \cite{zhou-etal-2020-scheduled, sun-etal-2022-alleviating} or entire encoder/decoder block layers \cite{sun-etal-2022-alleviating} during training, and employing the relaxed attention method \cite{lohrenz2021relaxed, lohrenz2022relaxed}. In our work, we achieve a similar effect of increased diversity by incorporating state, action region, and return prediction tasks alongside the primary action prediction task.

\section{Preliminaries}
\label{sec:headings}
We adopt the Markov decision process (MDP) framework to model our environment, denoted by $\mathcal{M}=$ $\langle\mathcal{S}, \mathcal{A}, p, P, R, \gamma\rangle$, where $\mathcal{S}$ and $\mathcal{A}$ represent the state and action spaces, respectively. The probability distribution over transitions is given by $P\left(s_{t+1} \mid s_t, a_t\right)$, while the reward function is defined as $R\left(s_t, a_t\right)$. The discount factor, $\gamma$, is used to weigh the importance of future rewards. The agent starts in an initial state $s_1$ sampled from the fixed distribution $p\left(s_1\right)$ and chooses an action $a_t \in \mathcal{A}$ from the state $s_t \in \mathcal{S}$ at each timestep $t$. The agent then transitions to a new state $s_{t+1}$ according to the probability distribution $P\left(\cdot \mid s_t, a_t\right)$. After each action, the agent receives a deterministic reward $r_t = R\left(s_t, a_t\right)$. We denote the quantization granularity of the action space as $b$ and the dimensionality of the action space as $d$.

\subsection{Setup and Notation}
Our goal is to model the offline RL problem as a sequence modeling problem, with the agent having access to a non-stationary training data distribution $\mathcal{T}$. We use $\tau$ to represent a trajectory and $|\tau|$ to denote its length. The return-to-go, which we refer to as 'return for short' throughout the paper, of a trajectory $\tau$ at timestep $t$ is defined as the sum of future rewards starting from that timestep, i.e., $g_t=\sum_{t^{\prime}=t}^T r_{t^{\prime}}$. The sequence of actions, action regions, states, and returns of trajectory $\tau$ are represented by $\mathbf{a}=\left(a_1, \ldots, a_{|\tau|}\right)$, $\mathbf{\bar{a}}=\left(\bar{a}_1, \ldots, \bar{a}_{|\tau|}\right)$, $\mathbf{s}=\left(s_1, \ldots, s_{|\tau|}\right)$, and $\mathbf{g}=\left(g_1, \ldots, g_{|\tau|}\right)$, respectively. We denote the quantization granularity of the action space as $b$ and the dimensionality of the action space as $d$.

\subsection{Decision Transformer}

The DT model is designed to process a trajectory $\tau$ as a sequence of three types of input tokens: returns, states, and actions. Formally, the trajectory $\tau$ is represented as follows: 
\begin{equation}
\label{eq:DT_traj}
\tau=\left(g_1, s_1, a_1, g_2, s_2, a_2, \ldots, g_{|\tau|}, s_{|\tau|}, a_{|\tau|}\right)
\end{equation}

The initial return $g_1$ is equivalent to the return of the trajectory. At each timestep $t$, DT utilizes the latest $K$ tokens to generate an action $a_t$. The value of $K$ is a hyperparameter referred to as the context length for the transformer, which may be shorter during evaluation than the one used during training. DT can learn either a deterministic or a stochastic policy $\pi_{DT}\left(a_t \mid \mathbf{g}_{-K, t}, \mathbf{s}_{-K, t}, \mathbf{a}_{-K, t-1}\right)$, where $\mathbf{s}_{-K, t}$ is the sequence of $K$ past states $\mathbf{s}_{\max (1, t-K+1): t}$ and similarly for $\mathbf{g}_{-K, t}$ whereas $\mathbf{a}_{-K, t-1}$ is the sequence of $k-1$ past actions. This creates an autoregressive model of order $K$. DT parameterizes the policy using a GPT architecture \cite{radford2018improving}, which applies a causal mask to enforce the autoregressive structure in the predicted action sequence.

In a standard probabilistic framework, the goal of DT is to learn a stochastic policy that maximizes the likelihood of the training data \cite{zheng2022online}. In this work, we specifically focus on continuous action spaces, which typically involve utilizing a multivariate Gaussian distribution with a diagonal covariance matrix \cite{haarnoja2018soft} to model the action distributions based on the previous returns, states, and actions. To represent the policy parameters, we use the symbol $\theta_{DT}$, and the policy that assigns a probability density to action $a_t$ at time step $t$ given the past $K$ state and return tokens and past $k-1$ action tokens $\mathbf{s}_{-K,t}$ and $\mathbf{g}_{-K,t}$ and $\mathbf{a}_{-k, t-1}$ is denoted by $\pi_{\theta_{DT}}(a_t|\mathbf{g}_{-K,t},\mathbf{s}_{-K,t}, \mathbf{a}_{-k-1, t-1})$. Formally, $\pi_{\theta_{DT}}$ can be expressed as follows:
\begin{equation}
\begin{aligned}
& \pi_{\theta_{DT}}\left(a_t \mid \mathbf{g}_{-K, t}, \mathbf{s}_{-K, t}, \mathbf{a}_{-k-1, t-1}\right) = \mathcal{N}\left(\mu_{\theta_{DT}}\left(\mathbf{g}_{-K, t}, \mathbf{s}_{-K, t}, \mathbf{a}_{-k-1, t-1}\right), \Sigma_{\theta_{DT}}\left(\mathbf{g}_{-K, t}, \mathbf{s}_{-K, t}, \mathbf{a}_{-k-1, t-1}\right)\right), \forall t
\end{aligned}
\end{equation}

DT learns the policy by minimizing the negative log-likelihood (NLL) loss, which can be expressed as follows:

\begin{equation}
\label{eq:DT_objective}
J_{DT}(\theta_{DT}) = -\frac{1}{K} \mathbb{E}_{(\mathbf{a}, \mathbf{s}, \mathbf{g}) \sim \mathcal{T}}\left[\sum_{k=1}^K \log \pi_{\theta_{DT}}\left(a_k \mid \mathbf{g}_{-K, k}, \mathbf{s}_{-K, k}, \mathbf{a}_{-K-1, k-1} \right)\right]
\end{equation}
Assuming that the covariance matrix $\Sigma_\theta$ is diagonal and that the variances are the same across all dimensions, the problem can be reduced to learning a deterministic policy with the standard $\ell_2$ loss, as done in the original DT \cite{chen2021decision}.

\section{Multi-Objective Decision Transformers}

In reinforcement learning contexts with continuous tokens, each token potentially exhibits unique distributions and dimensions. In such situations, employing the same linear projection for embedding identical token types, a common practice in Decision Transformer (DT), could potentially restrict the model's discriminative ability. Tokens with closely related distributions, albeit representing disparate environment aspects, may be assigned akin query (Q) and key (K) vectors. Let us consider tokens $x, x', y, y'$ of the same type. Assuming $Q(x) \approx Q(x')$ and $K(y) \approx K(y')$, where Q and K denote the query and key vectors respectively, the attention scores, which are computed via the dot product of query and key vectors, would yield $A(Q(x), K(y)) \approx A(Q(x'), K(y'))$. This observation implies that tokens of the same type contribute with similar relevance to the task of action prediction.

This phenomenon can be attributed to the training objective's narrow scope, which, in the current model, focuses solely on predicting the subsequent action. Consequently, the model's representational capacity might not be fully exploited, thereby affecting its ability to distinguish between tokens of the same type.

\subsection{Decision Transformer with State and Return Prediction}

In this work, we propose broadening the training objective to include not only action but also state and return predictions. Our hypothesis is that by diversifying the learning targets, we can guide the model towards learning more distinct representations for the input tokens. As a result, the DT model is expected to more effectively differentiate between tokens of the same type, even when their distributions are closely aligned. This approach aims to leverage the full representational capacity of the model, enhancing its performance in complex, multi-faceted reinforcement learning environments.

In alignment with the probabilistic formulation of DT which is delineated above, we introduce two probabilistic functions, $\rho_{\theta_{s}}$ and $\phi_{\theta_{g}}$, which respectively model the distribution of states distribution of returns. These functions are parameterized by $\theta_{s}$ and $\theta_{g}$ respectively, and they assign probability densities at the $t$-th time step, conditioned on the historical $k-1$ instances of state, action, and return tokens.

Specifically, the state distribution function $\rho_{\theta_{s}}$ is modeled using a multivariate Gaussian distribution with mean $\mu_{\theta_{s}}$ and a diagonal covariance matrix $\Sigma_{\theta_{s}}$, i.e. it can be expressed as follows:

\begin{equation}
\begin{aligned}
\rho_{\theta_{s}}\left(s_t \mid \mathbf{g}_{-K-1, t}, \mathbf{s}_{-K-1, t-1}, \mathbf{a}_{-K-1, t-1}\right) = \mathcal{N}\left(\mu_{\theta_{s}}\left(\mathbf{g}_{-K-1, t-1}, \mathbf{s}_{-K-1, t-1}, \mathbf{a}_{-K-1, t-1}\right), \right. \\
\left. \Sigma_{\theta_{s}}\left(\mathbf{g}_{-K-1, t-1}, \mathbf{s}_{-K-1, t-1}, \mathbf{a}_{-K-1, t-1}\right)\right), \forall t>1
\end{aligned}
\end{equation}

Similarly, the return distribution function $\phi_{\theta_{g}}$, is modeled as a univariate Gaussian with mean $\mu_{\theta_{g}}$ and standard deviation $\sigma_{\theta_{g}}$, i.e.

\begin{equation}
\begin{aligned}
\phi_{\theta_{g}}\left(g_t \mid \mathbf{g}_{-K-1, t-1}, \mathbf{s}_{-K-1, t-1}, \mathbf{a}_{-k-1, t-1}\right) = \mathcal{N}\left(\mu_{\theta_{s}}\left(\mathbf{g}_{-K-1, t-1}, \mathbf{s}_{-K-1, t-1}, \mathbf{a}_{-k-1, t-1}\right), \right. \\
\left.\sigma_{\theta_{g}}^2\left(\mathbf{g}_{-K-1, t-1}, \mathbf{s}_{-K-1, t-1}, \mathbf{a}_{-k-1, t-1}\right)\right), \forall t>1
\end{aligned}
\end{equation}

The loss functions associated with $\rho_{\theta_{s}}$ and $\phi_{\theta_{g}}$ are respectively defined as follows:
\begin{equation}
J_1(\theta_{s}) = -\frac{1}{K} \mathbb{E}_{(\mathbf{g}, \mathbf{s}, \mathbf{a}) \sim \mathcal{T}}\left[\sum_{k=1}^K \log \rho_{\theta_{s}}\left(\mathbf{s}_k \mid \mathbf{g}_{-K-1, k-1}, \mathbf{s}_{-K-1, k-1}, \mathbf{a}_{-k-1, k-1}\right) )\right]
\end{equation}

\begin{equation}
J_2(\theta_{g}) = -\frac{1}{K} \mathbb{E}_{(\mathbf{g}, \mathbf{s}, \mathbf{a}) \sim \mathcal{T}}\left[\sum_{k=1}^K \log \phi_{\theta_{g}}\left(\mathbf{g}_k \mid \mathbf{g}_{-K-1, k-1}, \mathbf{s}_{-K-1, k-1}, \mathbf{a}_{-K-1, k-1}\right))\right]
\end{equation}

These two objectives can be optimized jointly along with $J_{DT}$ in equation \ref{eq:DT_objective} in a multi-objective optimization (MOO) setting. Formally, let $\theta = \theta_{DT} \cup \theta_s \cup \theta_g$ be the set of the whole model parameters. The corresponding MOO problem is formulated as follows:

\begin{equation}
\label{eq:mo_dt_loss}
\theta^*=\underset{\boldsymbol{\theta}}{\arg \min } \lambda_0 J_{DT}(\theta) + \sum_{i=1}^2 \lambda_i J_i(\theta)
\end{equation}
where $\lambda_0, \lambda_1, \lambda_2$ are constants within the interval $[0,1]$, collectively satisfying the constraint $\lambda_0 + \lambda_1 + \lambda_2 = 1$. These parameters function as task-specific coefficients, determining the relative significance assigned to each task within the multi-objective optimization problem. As such, we introduce the designation "Multi-Objective Decision Transformer" for our model. 

This technique of using a weighted sum of the losses is known as linear scalarization. Despite theoretical constraints associated with this approach, such as potential loss of Pareto optimality and challenges navigating non-convex Pareto fronts, it offers satisfactory performance within the context of our problem. Indeed, one could consider integrating more sophisticated Pareto search techniques\cite{NeurIPS2018_Sener_Koltun, pmlr-v119-mahapatra20a} to enhance the multi-objective Decision Transformer. Yet, it is important to note that these advanced methodologies often carry a higher computational burden. Specifically, they typically necessitate solving a linear programming problem for each gradient update, which significantly prolongs the training duration. Moreover, these techniques often introduce additional hyperparameters, thereby increasing the complexity of model tuning. As such, while a potentially promising avenue for future research, this shift to more complex Pareto search techniques must be carefully balanced with considerations of computational efficiency and model simplicity.

\subsection{Trust Region Decision Transformer}

In the offline RL setting, trajectories are typically generated through random policies, leading to a high diversity of actions within each trajectory. This presents a potential issue when utilizing the MO-DT model, as the model may excessively focus on these diverse actions during the prediction of states, returns, and actions. Consequently, the model might become overly reliant on the potentially sub-optimal behavior policy, thereby limiting its generalization capabilities across different scenarios. In this section, we propose to smooth the trajectories with action regions.

\paragraph*{Trajectory Representation.}

We introduce the following modification to the trajectory representation presented in \ref{eq:DT_traj} by incorporating action regions as follows:

\begin{equation}
\tau=\left(g_1, s_1, \bar{a}_1, a_1, g_2, s_2, \bar{a}_2, a_2, \ldots, g_{|\tau|}, s_{|\tau|}, \bar{a}_{|\tau|}, a_{|\tau|}\right)
\end{equation}

Here, $\bar{a}_t$ represents the region of the action space in which $a_t$ is contained. To obtain these regions, we perform a coarse uniform discretization of the underlying action space. This technique has the advantage of retaining information about Euclidean distance in the original continuous action space, which may reflect the structure of the problem better than the training data distribution. Moreover, this discretization technique is straightforward to implement and does not require any additional assumptions about the underlying action distribution.

\paragraph{Action Region Representation and MO-TRDT Training.}

An intuitive way of representing action regions is through the use of one-hot encoding. However, this approach suffers from several limitations, most notably the curse of dimensionality. As the number of degrees of freedom increases, the number of action regions grows exponentially, leading to high-dimensional action region spaces. For instance, a 6-degree-of-freedom system like the HalfCheetah \cite{todorov2012mujoco} with a discretization granularity of 3, $a_i \in \{-1, 0, 1 \}$ for each joint, will have an action region space with a dimensionality of $3^6 = 729$. Training a transformer model in these settings would involve using a softmax distribution over all action regions, posing significant computational challenges that can hamper both training and inference, ultimately limiting scalability to problems with larger action spaces. Furthermore, the utilization of one-hot encodings disregards crucial information regarding the structure of the action domain, which can potentially hinder the ability to effectively solve certain problems.

To address the challenge of managing an explosion of the number of action regions in complex action spaces, we consider representing action regions as a vector of ordinal encodings \cite{cheng2008neural} per action dimension, instead of using one-hot encodings. Each dimension of the action space is represented by an ordinal encoding vector of length $b$. The final action region encoding is obtained by concatenating these ordinal encoding vectors for each dimension, resulting in an encoding vector with dimensionality $b \times d$. This straightforward approach retains information about the internal ordering between bins in each dimension, thereby preserving the structure of action regions. Furthermore, action regions can be efficiently classified using maximum likelihood estimation of a multivariate Bernouli distribution. This reduces the complexity of the classification problem, transitioning from an exponential growth, characterized by $b^d$, to a linear one, represented by $b \times d$. Formally, denoting the parameters used during action region prediction by $\theta_{\bar{a}}$, the likelihood function $\xi_{\theta_{\bar{a}}}$ that assigns a probability mass to action region $\bar{a}_t$ at time step $t$ given the past $K$ state and return tokens and past $k-1$ action and action region tokens can be expressed as follows:

\begin{equation}
\begin{aligned}
& \xi_{\theta_{\bar{a}}}\left(\mathbf{\bar{a}}_t \mid \mathbf{g}_{-K, t}, \mathbf{s}_{-K, t}, \mathbf{\bar{a}}_{-k-1, t-1}, \mathbf{a}_{-k-1, t-1}\right) = \prod_{j=1}^{b \times d} \mathcal{B}\left(p_{\theta_{\bar{a}}, j}\left(\mathbf{g}_{-K, t}, \mathbf{s}_{-K, t},  \mathbf{\bar{a}}_{-k-1, t-1}, \mathbf{a}_{-K-1, t-1}\right)\right), \forall t
\end{aligned}
\end{equation}
where $p_{\theta_{\bar{a}}, j}$ represents the probability of the $j$-th element in the binary action region vector, and $\mathcal{B}$ represents the Bernoulli probability mass function that models the distribution of each element in the action region vector. The product symbol indicates that we're considering the joint distribution of all elements in the action region vector $\mathbf{\bar{a}}_t$.

Subsequently, MO-TRDT learns $\xi_{\theta_{\bar{a}}}$  by minimizing the NLL loss, which can be expressed as follows: 

\begin{equation}
\label{eq:action_region_loss}
J_3(\theta_{\bar{a}}) = -\frac{1}{K} \mathbb{E}_{(\mathbf{g}, \mathbf{s}, \mathbf{\bar{a}}, \mathbf{a}) \sim \mathcal{T}}\left[\sum_{k=1}^K \log \xi_{\theta_{\bar{a}}}\left(\mathbf{\bar{a}}_k \mid \mathbf{g}_{-K, k}, \mathbf{s}_{-K, k}, \mathbf{\bar{a}}_{-k-1, k-1}, \mathbf{a}_{-k-1, k-1}\right) )\right]
\end{equation}

This objective function is equivalent to the binary cross-entropy loss. During inference, we employ a simple threshold-based approach, wherein values greater than 0.5 are classified as 1 and below as 0. This method is chosen over more sophisticated alternatives that enforce ordering, such as sequential prediction of entries per bin \cite{cheng2008neural} or utilizing factorized architectures \cite{tang2020discretizing}, which would add extra complexity to the algorithm in terms of both time and memory requirements.

Subsequently, the functions $\rho_{\theta_{s}}$, $\phi_{\theta_{g}}$, and $\pi_{\theta_{DT}}$ can be readily adapted to account for previous action regions. Importantly, we deliberately avoid conditioning the action $a_t$ at timestep $t$ on the corresponding action region $\bar{a}_t$. This strategic choice serves two purposes: first, it obviates the need to query the MO-TRDT for action region prediction during inference, thus averting potential delays in the inference process. Second, it reduces the likelihood of inaccuracies in subsequent predictions if the model fails to predict the optimal action region at a given timestep.

Given that functions $\rho_{\theta_{s}}, \phi_{\theta_{g}},$ and $\pi_{\theta_{DT}}$ have been adapted to account for action regions, the objective function in Eq. \ref{eq:mo_dt_loss} can be extended to incorporate the loss in Eq. \ref{eq:action_region_loss}, thereby enabling action region prediction. Formally, let $\theta_{MO-TRDT} = \theta_{DT} \cup \theta_s \cup \theta_g \cup \theta_{\bar{a}}$ represent the complete set of model parameters. The associated MOO problem is then articulated as:

\begin{equation}
\label{eq:MO-TRDT_loss}
\theta_{MO-TRDT}^*=\underset{\boldsymbol{\theta_{MO-TRDT}}}{\arg \min } \lambda_0 J_{DT}(\theta_{MO-TRDT}) + \sum_{i=1}^3 \lambda_i J_i(\theta_{MO-TRDT})
\end{equation}
where $\lambda_0, \lambda_1, \lambda_2, \lambda_3$ are constants within the interval $[0,1]$, satisfying the constraint $\lambda_0 + \lambda_1 + \lambda_2 + \lambda_3 = 1$.






\subsection{Inference of MO-DT and MO-TRDT}

In the evaluation phase of the MO-DT, we commence with a predetermined initial state $s_1$ and desired return $g_1$. The MO-DT produces the first action, denoted $a_1 = \mu_{\theta_{DT}}(\mathbf{s}_{1}, \mathbf{g}_1)$. Post execution of action $a_1$, the MO-DT predicts the subsequent return $g_2 = \phi_{\theta_g}(\mathbf{s}_1, \mathbf{g}_{1}, \mathbf{a}_1)$, a notable departure from the methodology employed by the standard DT where returns are not predicted during inference. Our empirical results demonstrate that this strategy yields more consistent results compared to a simplistic approach where the next return is derived by subtracting the acquired reward. Following this, the agent observes the subsequent state $s_{2}$ governed by the transition probability distribution $P\left(\cdot \mid s_1, a_1\right)$, and the MO-DT generates the next action $a_{2}$ predicated on $g_{1}, g_{2}$, $s_{1}, s_{2}$, and $a_{1}$ as inputs. This procedure persists until the termination of the episode.

A somewhat parallel pipeline is applied for the MO-TRDT, with the slight deviation that the associated action region is appended to the trajectory each time the model executes an action.


\section{Results and Analysis}

Our experiments are designed to comparatively evaluate our two proposed multi-objective decision transformers against prior offline RL methods. Specifically, we aim to understand how the incorporation of multi-objective optimization affects the attention mechanism in the Decision Transformer. Furthermore, we vary different components of our approach to demonstrate their significance.

\paragraph{Benchmark and Compared Baselines.}
We evaluate our algorithms, MO-DT and MO-TRDT, on the D4RL offline dataset for continuous control tasks \cite{fu2021d4rl}. Our experiments are conducted in the Gym domain \cite{brockman2016openai}, encompassing three environments (halfcheetah, hopper, walker2d), each with three levels (medium, medium-replay, medium-expert). All experiments utilize the 'v2' version of D4RL. For comparison, we consider our primary baseline, DT \cite{chen2021decision}, and the offline variant of the online decision transformer, ODT-O, proposed in \cite{zheng2022online}. This variant substitutes the deterministic policy of DT with a Gaussian policy. Results from the original papers of both methods are reported. For the experiments of 'medium-expert' for ODT-O are based on our reproduction following the hyper-paramters and code of the authors. We also compare against the two strongest dynamic programming-based state-of-the-art methods CQL\cite{kumar2020conservative} and IQL\cite{kumar2020conservative}. As the original paper for CQL reports performance on the 'V0' version of D4RL, which generally performs worse than 'V2', we refer to results reported in \cite{kostrikov2021offline}. Additionally, we compare with pure Behavioral Cloning (BC) and its variant, 10\%BC, which mimics the behavior of the top 10\%. For BC, we report results from \cite{kostrikov2021offline}, and for 10\%BC, we refer to \cite{chen2021decision}.

\begin{table}[t]
\centering
\caption{Averaged normalized scores on gym D4RL over 10 random seeds. Both of our methods outperform our primary baseline DT on almost all tasks, and matches or outperforms the best prior methods.}
\label{table:your_label}
\small 
\begin{tabularx}{\textwidth}{c|c|c|c|c|c|c|c|c}
\hline Dataset & BC & 10\% BC & CQL & IQL & DT & ODT-O& MO-DT(Ours) & MO-TRDT(Ours)\\
\hline halfcheetah-medium-v2 & 42.6 & 42.5 & 44.0 & \textbf{47.4} & 42.6 & 42.72& 43.34$\pm$ 3.34 & 43.20 $\pm$ 3.0\\
 hopper-medium-v2 & 52.9 & 56.9 & 58.5 & \textbf{66.3} & \textbf{67.6} & \textbf{66.95}& \textbf{67.45} $\pm$ 11.21 & \textbf{65.89} $\pm$ 8.38\\
 walker2d-medium-v2 & 75.3 & 75.0 & 72.5 & \textbf{78.3} & 74.0 & 72.19 & \textbf{78.56} $\pm$ 6.30 & 75.89 $\pm$ 7.26\\
halfcheetah-medium-replay-v2 & 36.6 & 40.6 & \textbf{45.5} & \textbf{44.2} & 36.6 & 39.99 & 40.93 $\pm$ 2.13 & 42.61 $\pm$ 2.0\\
 hopper-medium-replay-v2 & 18.1 & 75.9 & \textbf{95.0} & \textbf{94.7} & 82.7 & 86.64 & 79.43 $\pm$ 5.2 & \textbf{97.82} $\pm$ 2.3\\
 walker2d-medium-replay-v2 & 26.0 & 62.5 & 77.2 & 73.9 & 66.6 & 68.92 & 81.3 $\pm$ 9.72 & \textbf{86.94} $\pm$ 6.23\\
 halfcheetah-medium-expert-v2 & 55.2 & 92.9 & 91.6 & 86.7 & 86.8 & 88.76 & \textbf{94.57} $\pm$  2.78 & \textbf{94.37} $\pm$ 3.19\\
 hopper-medium-expert-v2 & 52.5 & \textbf{110.9} & 105.4 & 91.5 & 107.6 & 107.24 & \textbf{111.71} $\pm$ 1.52 & \textbf{111.53} $\pm$ 1.31\\
walker2d-medium-expert-v2 & \textbf{107.5} & \textbf{109.0} & \textbf{108.8} & \textbf{109.6} & \textbf{108.1} & \textbf{108.65} & \textbf{108.08} $\pm$ 0.54 &  \textbf{108.52} $\pm$ 0.45\\
\hline 
Total & 466.7 & 666.2 & 698.5 & 692.4 & 672.6 & 674.06 & 705.37 & \textbf{726.77}
\\
\hline
\end{tabularx}
\end{table}

\subsection{Investigating the learned attention patterns}
\begin{figure}[!h]
    \centering
    \begin{subfigure}[b]{0.27\textwidth}
        \centering
        \includegraphics[width=\textwidth]{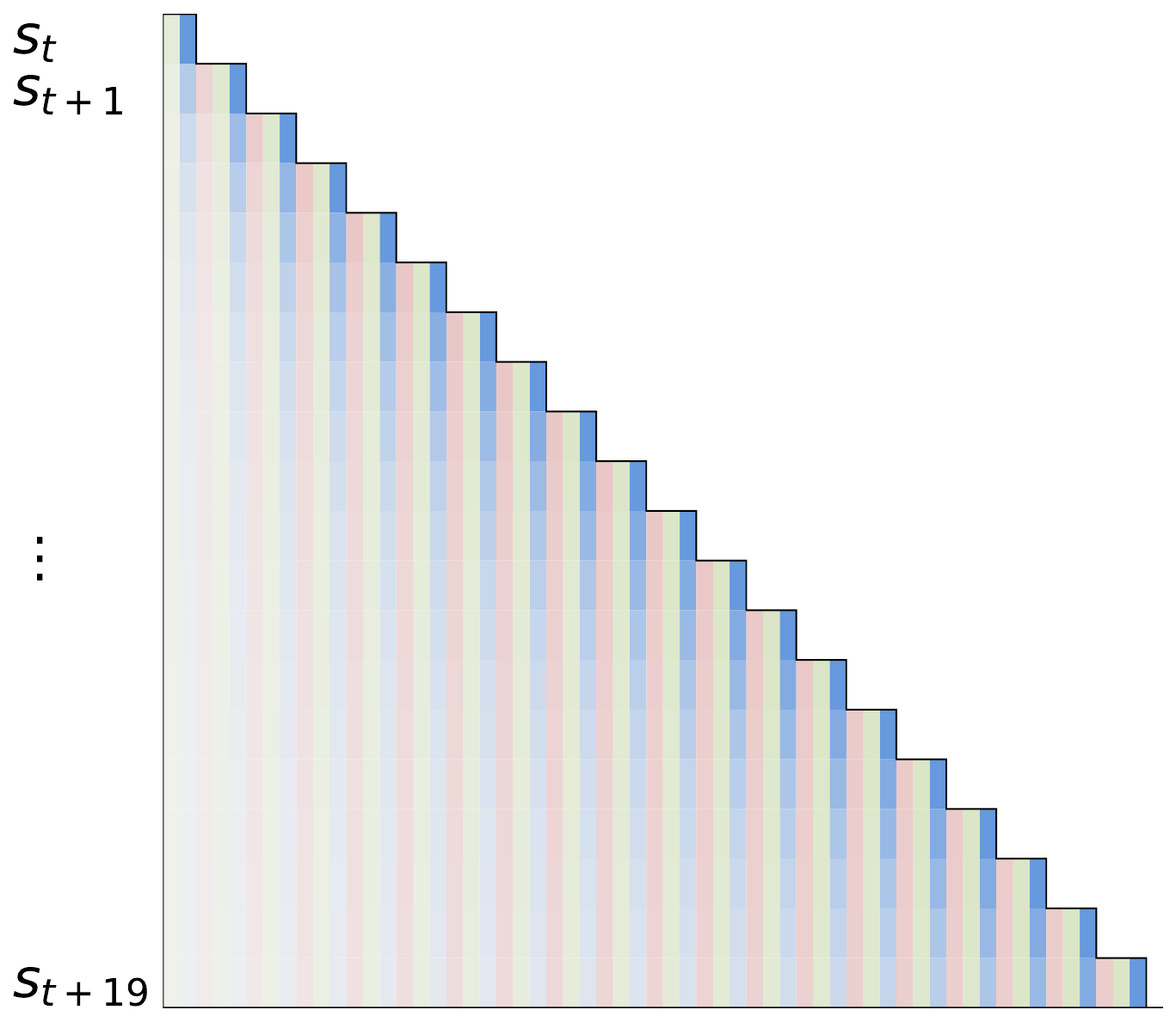}
        \label{fig:subfig1}
    \end{subfigure}
    \hfill
    \begin{subfigure}[b]{0.23\textwidth}
        \centering
        \includegraphics[width=\textwidth]{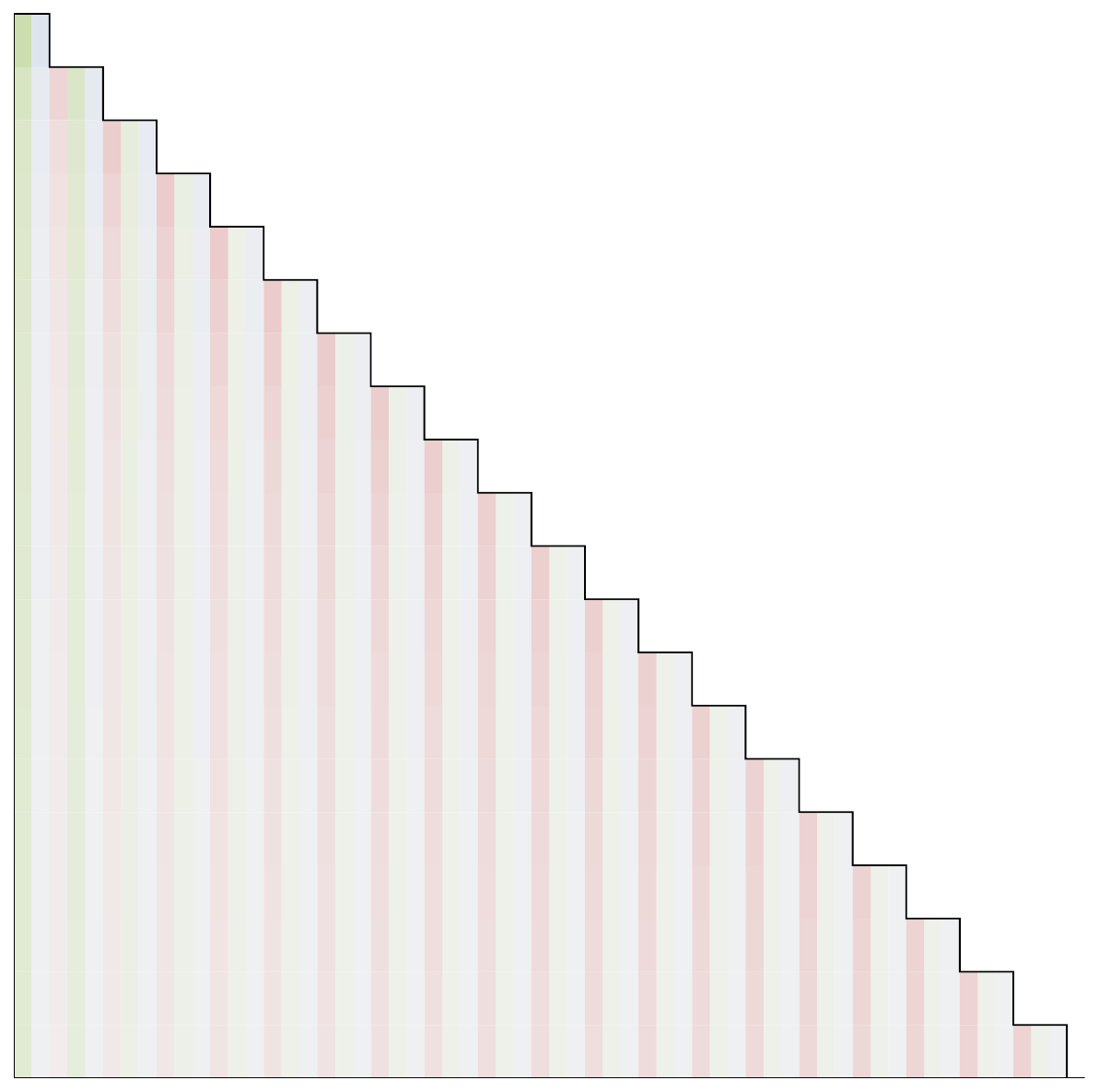}
        \label{fig:subfig2}
    \end{subfigure}
    \hfill
    \begin{subfigure}[b]{0.23\textwidth}
        \centering
        \includegraphics[width=\textwidth]{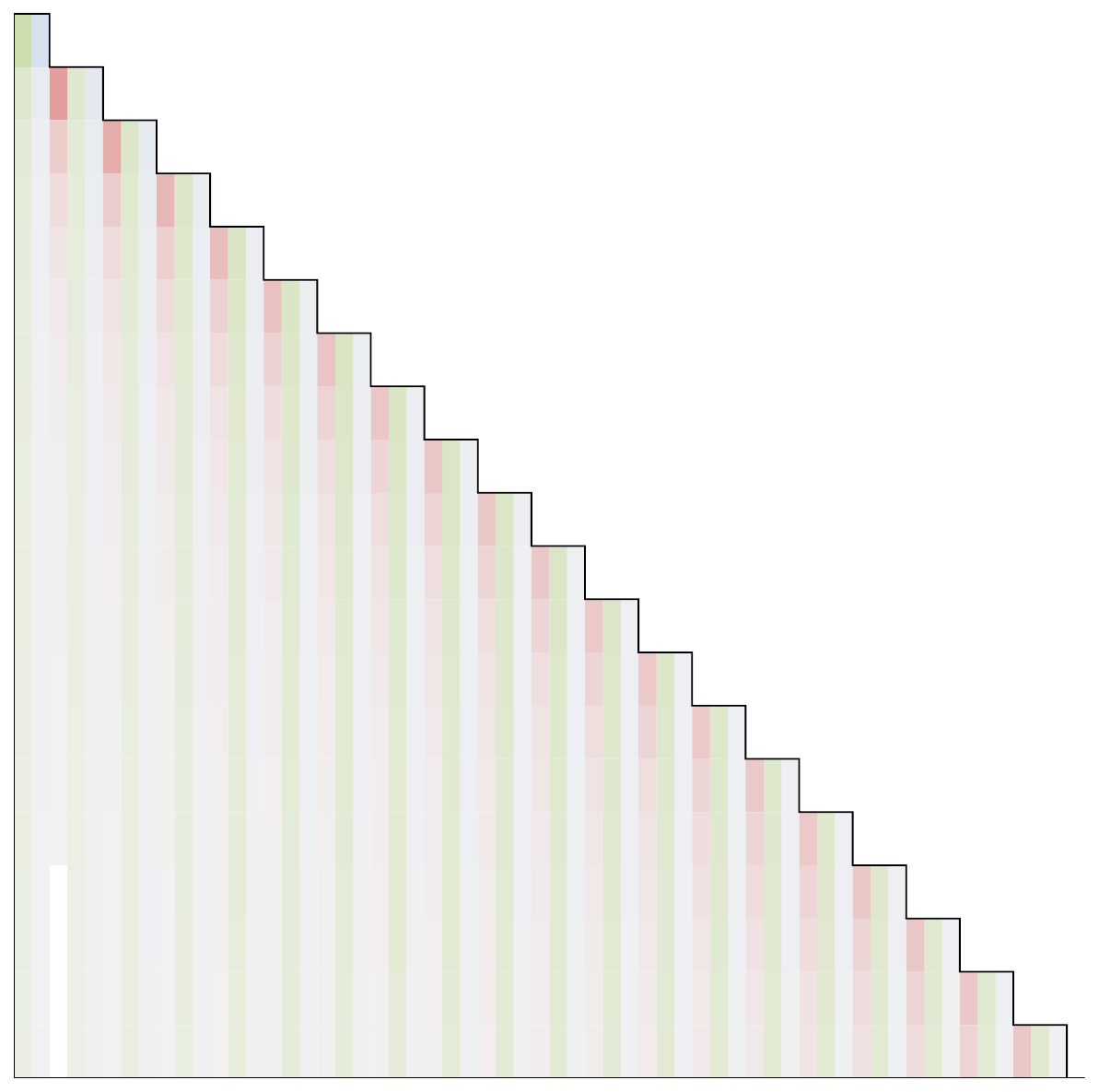}
        \label{fig:subfig3}
    \end{subfigure}
    \hfill
    \begin{subfigure}[b]{0.23\textwidth}
        \centering
        \includegraphics[width=\textwidth]{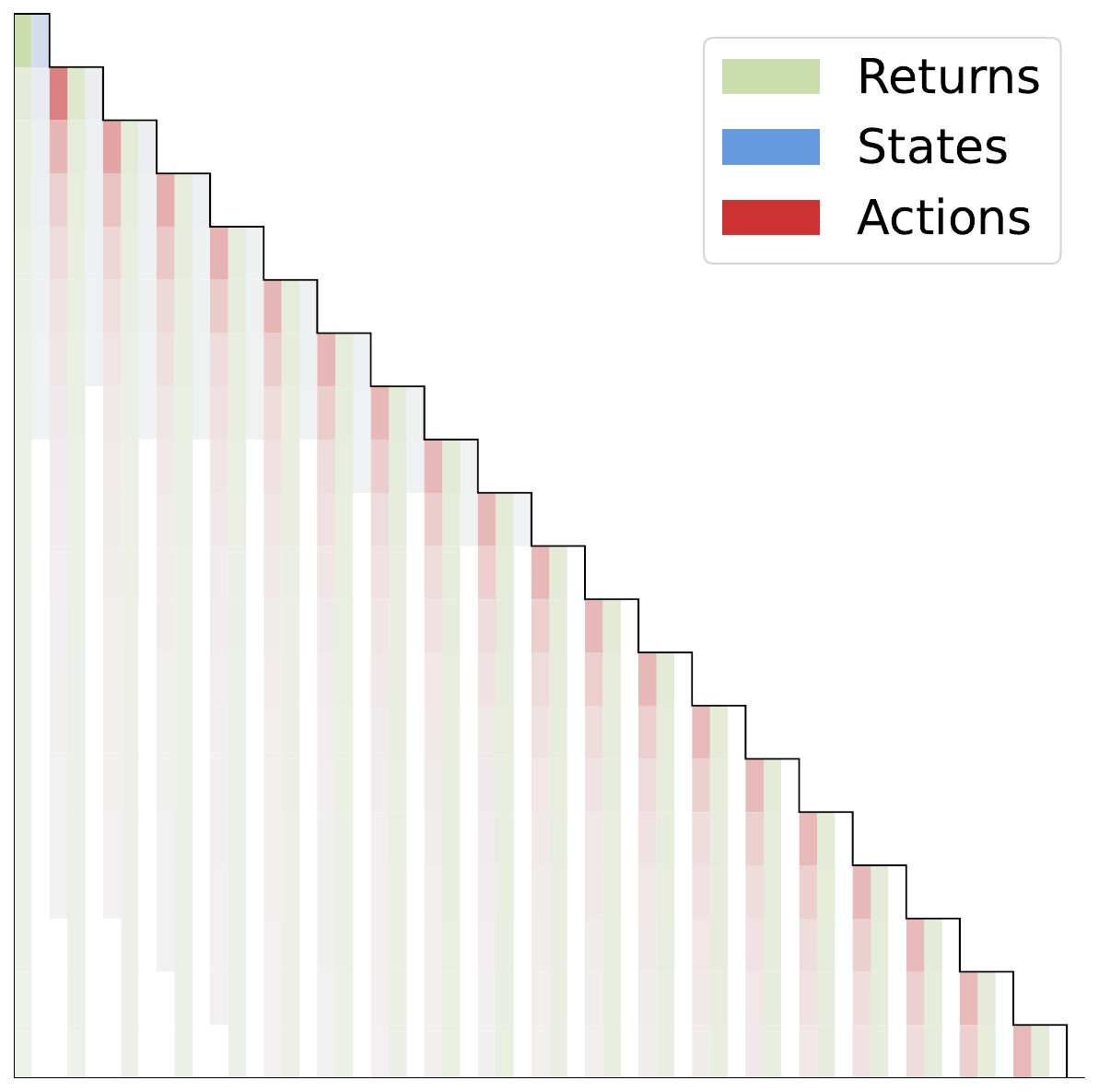}
        \label{fig:subfig4}
    \end{subfigure}
    \caption{Block-wise attention patterns for MO-DT during action prediction on Walker2d; unlike DT, MO-DT demonstrates a diversified focus on different tokens per block, indicative of a more efficient utilization of the transformer's attention mechanism. 
    }
    \label{fig:attention_maps_mo_dt}
\end{figure}

\begin{figure}[!h]
    \centering
    \begin{subfigure}[b]{0.27\textwidth}
        \centering
        \includegraphics[width=\textwidth]{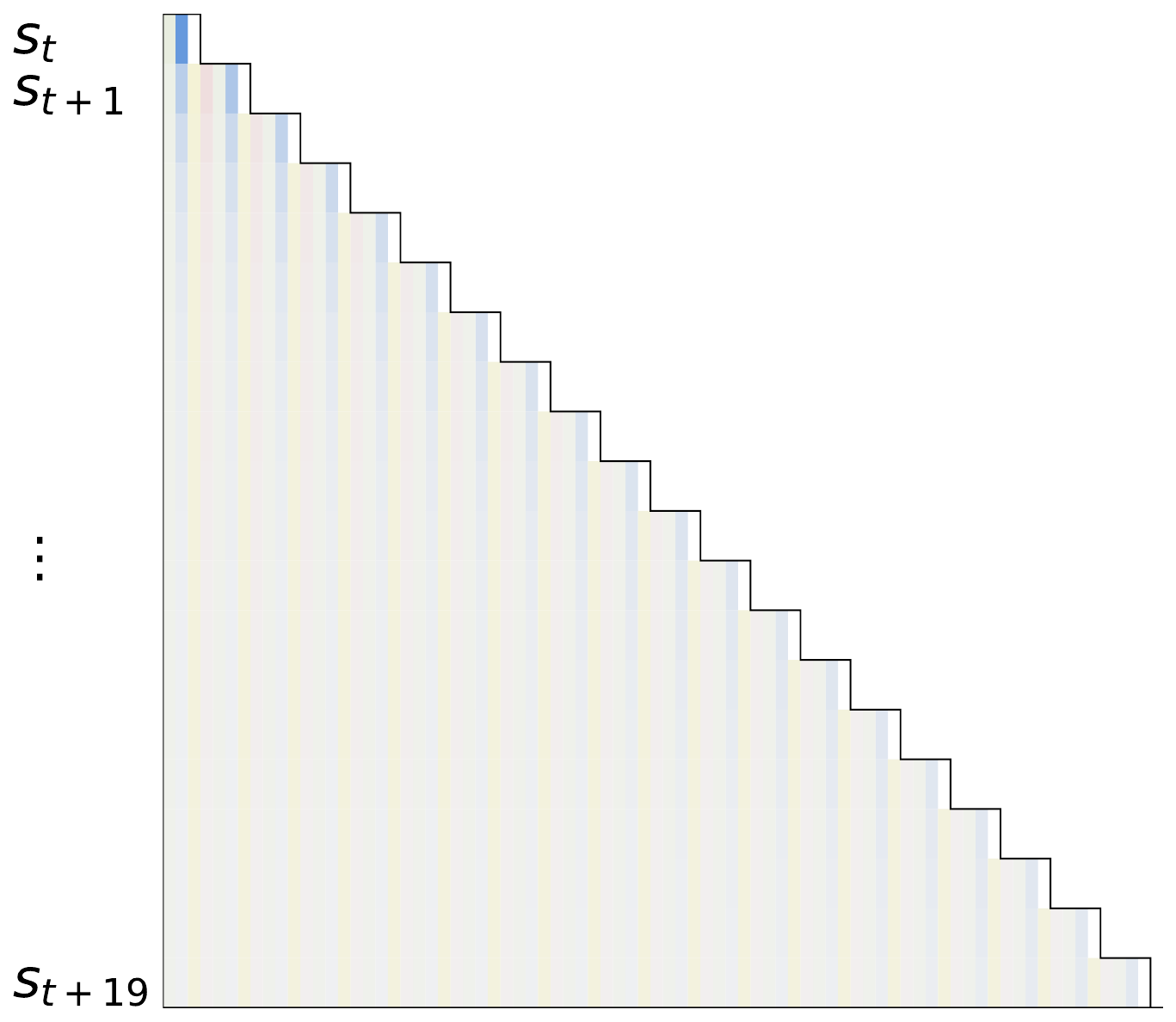}
        \label{fig:subfig1}
    \end{subfigure}
    \hfill
    \begin{subfigure}[b]{0.23\textwidth}
        \centering
        \includegraphics[width=\textwidth]{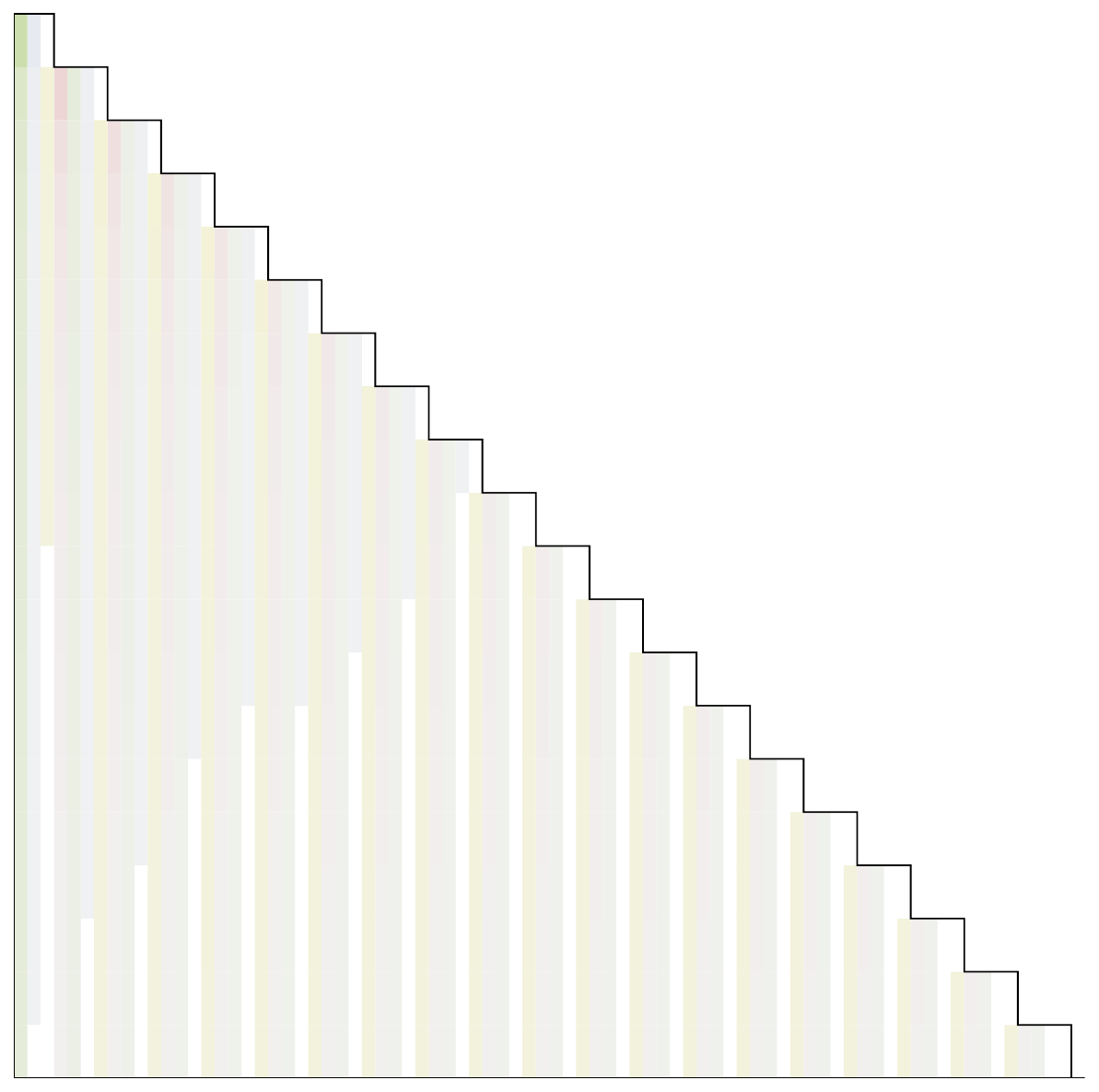}
        \label{fig:subfig2}
    \end{subfigure}
    \hfill
    \begin{subfigure}[b]{0.23\textwidth}
        \centering
        \includegraphics[width=\textwidth]{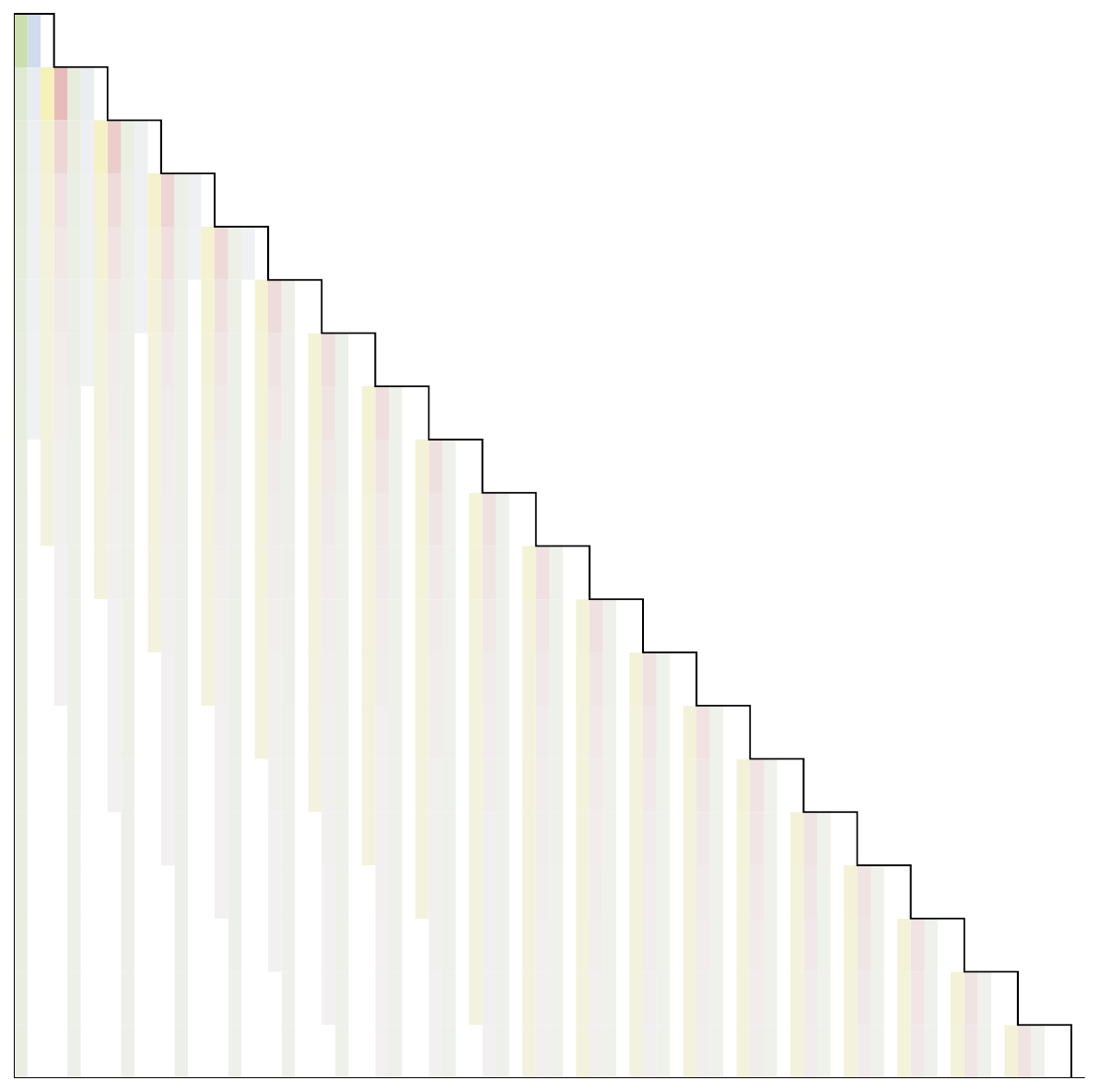}
        \label{fig:subfig3}
    \end{subfigure}
    \hfill
    \begin{subfigure}[b]{0.23\textwidth}
        \centering
        \includegraphics[width=\textwidth]{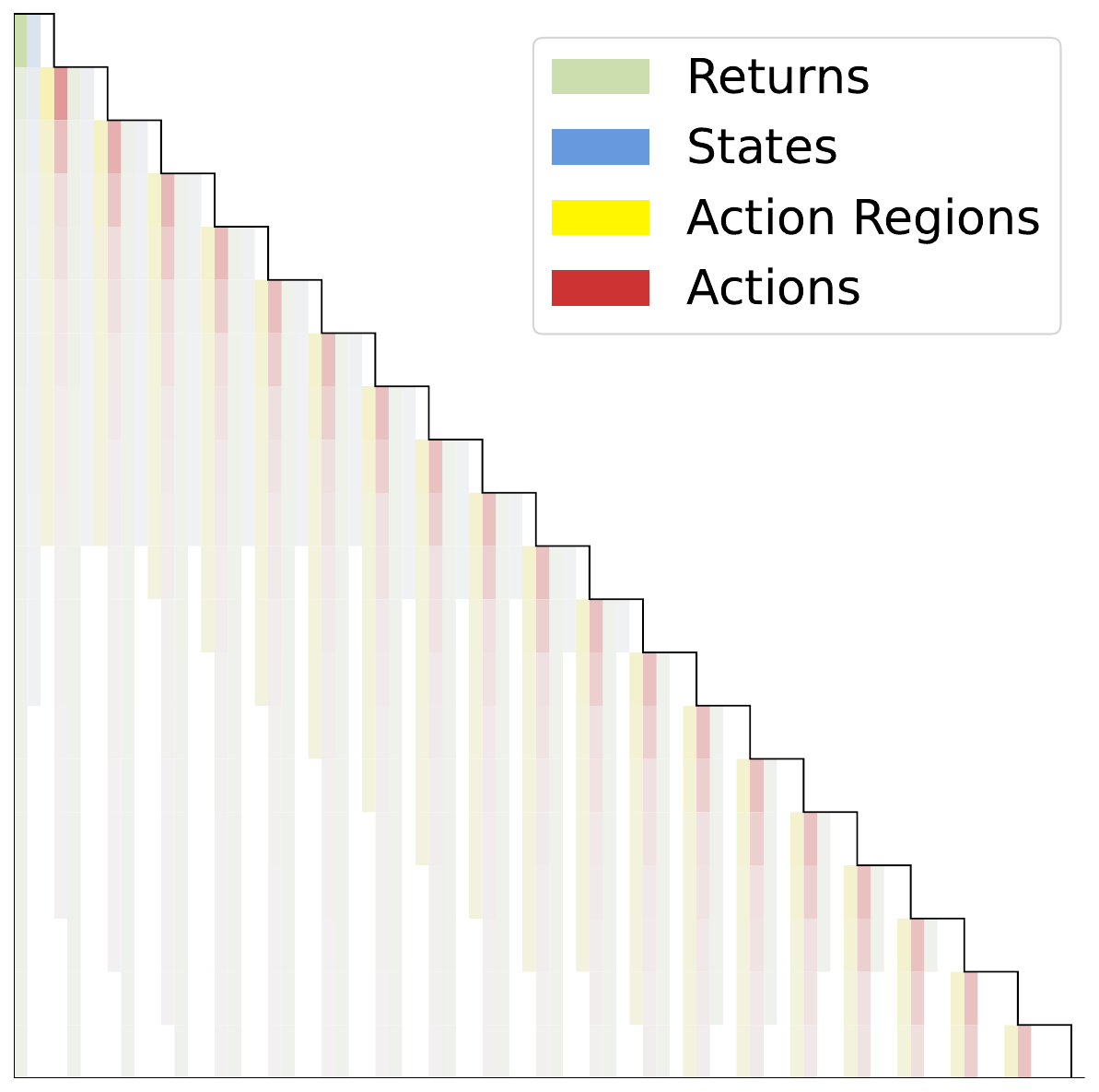}
        \label{fig:subfig4}
    \end{subfigure}
    \caption{Block-wise attention patterns for MO-TRDT during action prediction on Walker2d; the inclusion of action regions appears to promote a more localized distribution of attention, potentially indicating an enhanced capacity of the transformer model to distinguish between tokens of the same type. Notably, the attention at later blocks seems to be allocated between both actions and action regions.}
    \label{fig:attention_maps_MO-TRDT}
\end{figure}

In Figures \ref{fig:attention_maps_mo_dt} and \ref{fig:attention_maps_MO-TRDT}, we present attention maps for MO-DT and MO-TRDT models, respectively. Firstly, we discern that both models prompt the decision transformer to focus selectively on distinct tokens per block, with the attention appearing sparser for the MO-TRDT case, contrasting the uniform attention patterns displayed by DT as shown in Figure \ref{fig:attention_maps_dt}. This implies that multi-objective optimization in our configuration potentially mirrors the effects of contemporary transformer regularization techniques. Secondly, we note that later blocks of the transformer model predominantly engage with high-level representations of action tokens over return or state tokens. This observation aligns with the findings by \cite{tt} for the Trajectory Transformer. Lastly, for MO-TRDT, attention in subsequent blocks is divided between action regions and action tokens, hinting at a decreased dependency on the behavioral policy. Despite the derivation of action regions from the behavior policy, their coarse nature (we use $b=3$ across all experiments) leads to their presence in a wider array of trajectories and contexts, possibly enhancing the DT model's capacity to effectively integrate different trajectory segments. 

\begin{figure}[h!]
    \centering
    \begin{subfigure}[b]{0.45\textwidth}
        \includegraphics[width=\textwidth]{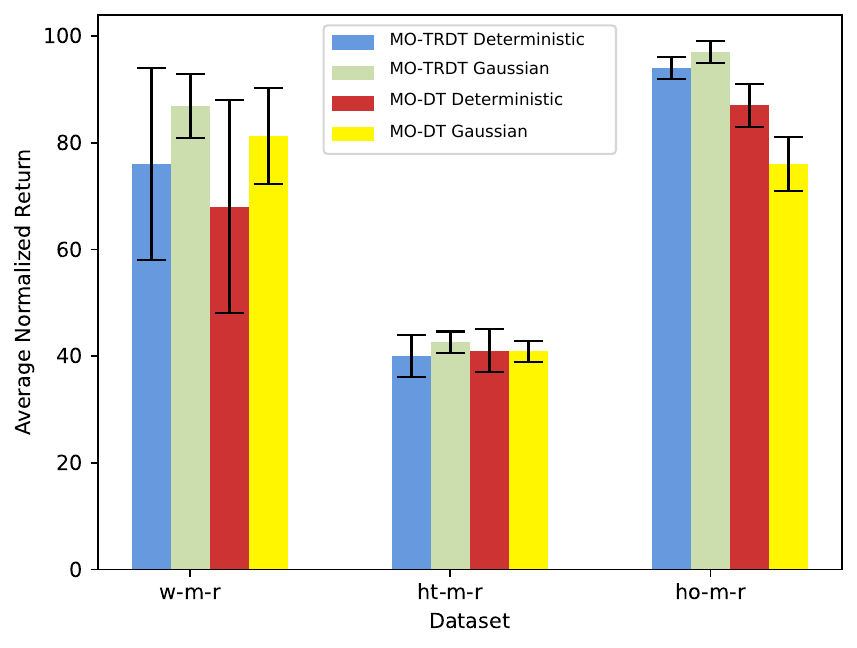}
        \caption{The effect of gaussion prediction heads.}
        \label{fig:gaussian_effect}
    \end{subfigure}
    \hfill
    \begin{subfigure}[b]{0.45\textwidth}
        \includegraphics[width=\textwidth]{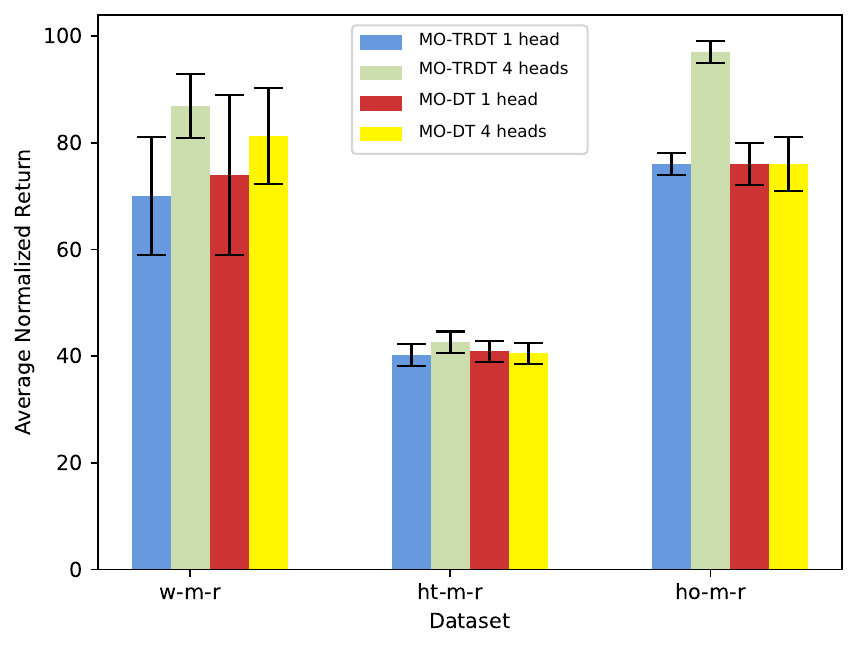}
        \caption{The effect of multi-head attention.}
        \label{fig:attention_effect}
    \end{subfigure}
    \caption{Ablation studies showing the effect of each design choice in our proposed methods across w-m-r (walker2d-medium-replay-v2), ht-m-r (halfcheetah-medium-replay), and ho-m-r (hopper-medium-replay) datasets.}
    \label{fig:ablation_studies}
\end{figure}

\paragraph*{Gaussian vs Deterministic Attention Heads.}
To substantiate our choice of Gaussian prediction heads for states and returns over linear prediction heads (trained using mean squared error minimization), we compared MO-DT and MO-TRDT both with and without Gaussian prediction heads in figure \ref{fig:gaussian_effect}. For simpler environment dynamics, as in the Hopper environment (action dimensionality 3, state dimensionality 6), deterministic prediction heads perform on par with their Gaussian counterparts. However, in the face of complex dynamics, deterministic heads yield high variance within model predictions, an established observation in RL literature \cite{haarnoja2018soft}.

\paragraph*{Importance of Multi-Head Attention.}
In Figure \ref{fig:attention_effect}, our aim is to evaluate the extent to which our model leverages the inherent capabilities of the transformer architecture. We observe that employing a singular attention head detrimentally affects the performance of both models. The common purpose of multiple attention heads is to facilitate simultaneous attention to diverse representation subspaces at distinct positions. Given this, the performance degradation with a single attention head implies that our model is effectively utilizing multiple representation subspaces in parallel, thus capitalizing on the transformer architecture's potential.

\section{Conclusion}
We proposed Multi-objective Decision Transformer (MO-DT), a Decision Transformer variant that effectively harnesses the transformer architecture by integrating state and return prediction in a multi-objective setting. We further extended MO-DT to Trust Region Decision Transformer, which incorporates action region prediction, decreasing dependency on the behavior policy and enhancing generalization. To our understanding, this work pioneers the exploration of state and return prediction's impact on the Decision Transformer method. Lastly, we demonstrated that our approaches markedly surpass the original Decision Transformer and meet or exceed prior top-performing methods on numerous D4RL benchmark locomotion tasks.


\bibliographystyle{unsrt}  
\bibliography{templateArxiv}  
\newpage
\section*{Appendices}
\appendix

\section{Hyperparameters of MO-DT and MO-TRDT}

In this section, we detail the optimal architectural and hyperparameter settings for MO-DT and MO-TRDT. Our chosen model is half the size of the ODT-O model presented in \cite{zheng2022online}. Specifically, our model employs a Transformer structure comprising four layers, each equipped with four attention heads. The embedding size is set to 256, deviating from the 512-dimension embedding used in ODT-O. Our empirical studies revealed no significant performance gain by increasing the embedding size to 512.

For both models, we used the LAMB optimizer \cite{you2020large} to optimize the model parameters. In terms of scalarization coefficients, uniform values have consistently yielded superior results across all conducted experiments with both models. Finally, consistent with the findings from \cite{zheng2022online}, our models do not incorporate positional embeddings. We found this approach to yield superior results compared to configurations that include them. For MO-TRDT, we used an action region granularity $b=3$ for all experiments. The full list of hyper-parameters is summarized in \ref{tab:hyperparameters}.

\begin{table}[h]
\centering
\begin{tabular}{l l}
\hline
\textbf{Hyperparameter} & \textbf{Value} \\
\hline
Number of bins (MO-TRDT) & $b=3$\\

Linear scalarization coefficients (MO-DT) & $\lambda_0=\lambda_1=\lambda_2=1/3$\\

Linear scalarization coefficients (MO-TRDT) & $\lambda_0=\lambda_1=\lambda_2=\lambda_3=1/4$\\

Number of layers  & 4 \\

Number of attention heads  & 4 \\

Embedding dimension & 256 \\

Context Length K & 20 \\

Dropout  & 0.1 \\

Nonlinearity function & ReLU \\

Batch size & 256 \\

Learning rate & 0.0001 \\

Weight decay & 0.001 \\

Gradient norm clip  & 0.25 \\

positional embedding & No \\

Learning rate warmup & linear warmup for $10^4$ training steps \\

Total number of updates & $10^5$ \\
\hline
\end{tabular}
\caption{Summary of Hyperparameters used to train MO-DT and MO-TRDT}
\label{tab:hyperparameters}
\end{table}

In terms of the initial return prompt for both models, empirical evidence indicated that initializing with a return value twice that of the expert's return yielded superior results across all tested datasets, with the exception of the HalfCheetah dataset. For this particular dataset, the use of the expert return proved most effective. The exact return values are summarized in table \ref{tab:return_vals}
\begin{table}[h]
\centering
\begin{tabular}{l l}
\hline
\textbf{Domain} & \textbf{Initial Return Value} \\
\hline
Halcheetah & 12000\\

Walker2d& 10000\\

Hopper& 7200\\
\hline
\end{tabular}
\caption{Initial Return value for each environment}
\label{tab:return_vals}
\end{table}

\end{document}